\documentclass[conference]{IEEEtran}
\IEEEoverridecommandlockouts
\usepackage{cite}
\usepackage{amsmath,amssymb,amsfonts}
\usepackage{algorithmic}
\usepackage{graphicx}
\usepackage{textcomp}
\usepackage{xcolor}
\usepackage{multirow}
\usepackage{subcaption}

\begin{document}

\title{CEC-CNN: A Consecutive Expansion-Contraction Convolutional Network for Very Small Resolution Medical Image Classification\\
}

\author{
\IEEEauthorblockN{
Ioannis Vezakis\IEEEauthorrefmark{1}, 
Antonios Vezakis\IEEEauthorrefmark{2},
Sofia Gourtsoyianni\IEEEauthorrefmark{2},\\
Vassilis Koutoulidis\IEEEauthorrefmark{2},
George K. Matsopoulos\IEEEauthorrefmark{1} and 
Dimitrios Koutsouris\IEEEauthorrefmark{1}}
\\
\IEEEauthorblockA{\IEEEauthorrefmark{1}Biomedical Engineering Laboratory, School of Electrical and Computer Engineering\\National Technical University of Athens, Greece}
\\
\IEEEauthorblockA{\IEEEauthorrefmark{2}Aretaieion Hospital, Medical School\\National and Kapodistrian University of Athens, Greece}
}

\maketitle

\begin{abstract}
Deep Convolutional Neural Networks (CNNs) for image classification successively alternate convolutions and downsampling operations, such as pooling layers or strided convolutions, resulting in lower resolution features the deeper the network gets. 
These downsampling operations save computational resources and provide some translational invariance as well as a bigger receptive field at the next layers. 
However, an inherent side-effect of this is that high-level features, produced at the deep end of the network, are always captured in low resolution feature maps. 
The inverse is also true, as shallow layers always contain small scale features. 
In biomedical image analysis engineers are often tasked with classifying very small image patches which carry only a limited amount of information. 
By their nature, these patches may not even contain objects, with the classification depending instead on the detection of subtle underlying patterns with an unknown scale in the image's texture.
In these cases every bit of information is valuable; thus, it is important to extract the maximum number of informative features possible. 
Driven by these considerations, we introduce a new CNN architecture which preserves multi-scale features from deep, intermediate, and shallow layers by utilizing skip connections along with consecutive contractions and expansions of the feature maps. 
Using a dataset of very low resolution patches from Pancreatic Ductal Adenocarcinoma (PDAC) CT scans we demonstrate that our network can outperform current state of the art models. 
\end{abstract}


\section{Introduction}
Extracting complex, high-order features from images using Convolutional Neural Networks (CNNs) requires a sufficiently deep network architecture, the reason being that as a general rule, shallow layers learn to respond to simple characteristics such as corners and edges, while deep layers are able to find more complex patterns, shapes, and high-level semantic information. 
It is thus expected that deep CNNs perform better than their shallower counterparts in increasingly complex problems. 
Indeed, from AlexNet \cite{krizhevskyImageNetClassificationDeep2017} to VGG \cite{simonyanVeryDeepConvolutional2015} and ResNet \cite{heDeepResidualLearning2015}, deeper networks have been corresponding to a clear performance increase. 

Deep CNNs consist of stacks of convolutional layers, separated by downsampling operations. 
These operations are most often pooling layers (e.g. max-pooling) or strided convolutions. 
They effectively summarize neighborhoods of feature maps, achieving translation invariance and a bigger receptive field at subsequent convolutional layers \cite{yamashitaConvolutionalNeuralNetworks2018}. 
As a result of these downsampling operations, high-level features, produced at the deep end of the network, will always be captured in low resolution feature maps.  
Similarly, fine-grained features will always be constrained to shallow layers. 
This is a major issue in applications where deep, fine-grained features need to be captured, e.g. for object detection. 
U-Net famously worked around this problem by using down-sampling layers to capture an image's context first, then up-sampling to achieve precise localization \cite{ronnebergerUNetConvolutionalNetworks2015}. 

Consider small image patches cropped from cancerous and non-cancerous regions of CT scans. 
In organs such as the pancreas, their differences can be so subtle that they become almost indistinguishable to the naked eye.
The complexity and size of the underlying features to be learnt is unknown. 
With a typical CNN, complex patterns captured by deep features could be lost if they were small in scale to begin with.
On the other hand, shallow features might be too localized when they actually need to be calculated over a large portion of the image. 
To this end, we design a network architecture that preserves shallow, intermediate, and deep feature maps of various resolutions until the very end of the network, where they are finally combined, refined, and classified. 
This is achieved by using consecutive expansion (upsampling) and contraction (downsampling) operations between convolutional layers, as well as skip connections between feature maps of the same resolution size. 

\section{Background}

Through the development of new network architectures, CNNs are becoming increasingly more effective for image classification tasks \cite{krizhevskyImageNetClassificationDeep2017} \cite{szegedyGoingDeeperConvolutions2015} \cite{simonyanVeryDeepConvolutional2015} \cite{heDeepResidualLearning2015} \cite{tanEfficientNetRethinkingModel2020}.
In general, the deeper a network is, i.e. the more consecutive convolutional layers it has, the better it is expected to perform. 

Each convolutional layer produces a feature map, whose pixel values correspond to a response to a particular feature in the input image. 
By summarizing a feature map's regions into single pixels (i.e. downsampling the map), some translational invariance is achieved at the cost of reducing spatial information. 
For example, max-pooling is a popular operation of this type, which, for a given region of the map, selects the maximum pixel value and discards the rest. 
Even though discarding data is sub-optimal, pooling operations are particularly useful for an additional reason; they multiplicatively increase the receptive field of subsequent convolutions \cite{luoUnderstandingEffectiveReceptive2016}. 

A Receptive Field (RF) is the region of the input image that influences the calculation of a single element in a feature map.
Feature maps produced with a high RF are influenced by larger, more general features in the input image, while a smaller RF translates to finer, more localized, features. 
In theory, the RF increases linearly as more convolutional layers are stacked. 
In practice, the Effective Receptive Field (ERF) occupies only a fraction of the theoretical RF, and pooling is useful because it is much more effective in enlarging it than convolutional layers \cite{luoUnderstandingEffectiveReceptive2016}. 
Thus, in a typical CNN, shallow feature maps correspond to small ERFs, while deep feature maps correspond to ERFs that occupy almost the entire image. 

The architecture of current CNNs is biased, in the sense that they are made to perform well on datasets such as the ImageNet \cite{dengImageNetLargescaleHierarchical2009}, where the object of interest occupies a significant portion of the input image. 
In these cases, a large ERF at the last feature maps is particularly useful, as each deep feature produced is influenced by the broader context of the image. 
When the objects in question are small however, large ERFs are no longer the answer \cite{zhangSmallObjectDetection2020} \cite{sunMultipleReceptiveFields}. 
Features are likely to not get picked up when there is a big mismatch between the RF's size and theirs. 

In very small images we do not have the luxury of making assumptions regarding our data. 
Especially when there is no particular object of interest as the image's overall texture needs to be classified, there is no guarantee on the scale, spatial location, or complexity of the features. 
It is our assumption that a multi-scale approach, based on multiple ERFs at various depths, can provide better results than traditional CNNs on small signal-to-noise images.

\section{Related Work}

Deep features used for image classification are not suitable for tasks targeting smaller regions of an image, e.g. object detection or segmentation, due to their low resolution. 
To this end, networks for these tasks often utilize upsampling methods to increase the resolution of their deep feature maps. 
For example, U-Net, named after it's U-shaped architecture and used for segmentation tasks, consists of a contracting path and an expansive path \cite{ronnebergerUNetConvolutionalNetworks2015}. 
The contractive path is composed of blocks of convolutional layers and max-pooling operations, doubling the amount of features in every step. 
The expansive path is more complex; it consists of an upsampling operation, then a convolution that halves the amount of features, a concatenation with the high-resolution feature maps from the contracting path, and two more convolutions to refine the concatenated features. 
As such, U-Net re-uses shallow, high resolution features, by combining them with the upsampled deep features of the same resolution. 

The stacked hourglass is another CNN architecture that makes use of upsampling mechanisms \cite{newellStackedHourglassNetworks2016}. 
It consists of stacked expansion-contraction modules.  Convolutional and max-pooling layers produce features of very low resolution first. 
Then, the module upsamples the low-res feature maps and combines them with features produced at the downsampling phase. 
In contrast to U-Net, this operation is done repeatedly, resulting in an architecture whose shape resembles that of stacked hourglasses. 

MSDNet \cite{huangMultiScaleDenseNetworks2018} strives to preserve multi-scale feature maps through the network. 
Each layer produces features of all scales, with each consecutive feature map being produced using dense connections from the same and higher-resolution features. 
Multiple classifiers present throughout the network allow the model to select the network depth, and thus the amount of computation for a particular input. 
In contrast to previous architectures, MSDNet does not contain any upsampling mechanisms for high resolution deep feature maps. 

FishNet is a CNN framework that builds upon expansion-contraction architectures with the goal of offering a unified approach over pixel-level and region-level tasks \cite{sunFishNetVersatileBackbone2019}. 
Inspired by networks such as U-Net, Stacked Hourglass, and MSDNet, it is built upon the concept of preserve and refine - it preserves features of different depths and uses them to refine each other.
Its architecture resembles the shape of a fish, as it consists of 3 paths: a contracting path (tail), an expansive path (body), and a final contracting path (the head). 
The tail and body resemble U-Net's structure, as the tail consists of a series of convolutional and max-pooling blocks, while the body consists of upsampling blocks and more convolutional layers.
Features from the shallow layers are concatenated with features of the same resolution in the deeper layers. 
It is important to mention that any skip connections in FishNet's architecture consist solely of identity mapping, concatenation, and max-pooling. 
Consider networks such as ResNet, which also use skip connections, but these skip connections sometimes include a convolutional layer to match the number of input and output channels. 
FishNet's careful design to not use convolutions in the skip connections enables the direct backpropagation (direct BP) of the gradient from deep to shallow layers. 
This tackles the problem of degradation, which can result in the accuracy getting saturated and then starting to degrade rapidly as the network depth keeps increasing. 

\section{Network Design} \label{design}

Our network design builds upon FishNet, exploiting its architecture to go even deeper and preserve all available features in the process. It is made up of 5, alternating, expansion and contraction parts. 

The first part of the network (hereafter referred to as CEC-CNN due to its Consecutive Expansion-Contraction architecture) is made purely of Bottleneck Residual Blocks (ResBlocks). 
They have three convolutional layers, each followed by a batch normalization. 
The first two are additionally followed by a ReLU operation and have kernel sizes of 1 and 3 respectively, while the last one, which also has a kernel size of 1, is followed by an element-wise addition of its output with the block's original input. 
All convolutions have a stride of 1, and their output filter number matches the input's. 
In this way we ensure that the skip connection from the input to the output is always a simple identity mapping in order to facilitate direct backpropagation of the gradient to the shallow layers. 
The only exception is the first convolutional layer of ResBlocks that need to downsample the input feature map, where a stride of 2 is used, and the element-wise addition operation is omitted. 

\begin{figure*}[htbp]
\centerline{\includegraphics[width=0.85\textwidth]{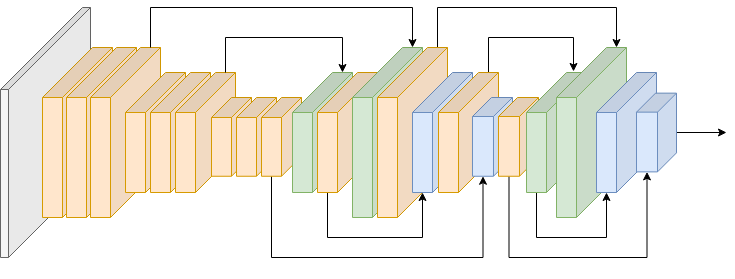}}
\caption{The CEC-CNN architecture. Bottleneck ResBlocks are depicted in yellow, U-Blocks in green, and D-Blocks in blue. Input image is in grey. The arrows indicate concatenation of the feature maps at their source with the feature maps at their destination. Each downsampling layer doubles the amount of feature maps, while each upsampling layer halves it.}
\label{fig}
\end{figure*}

The next parts combine Upsampling (U-Blocks), Downsampling (D-Blocks), and ResBlocks. 
The U-Blocks and D-Blocks are similar to fishnet's UR and DR blocks, but the order of operations is flipped. 
Instead of upsampling, concatenating, and refining the input, the U-Block refines, upsamples, then concatenates the input. 
This offers two advantages: i) it decreases the number of the input feature maps, lowering the amount of parameters, ii) it allows the shallow feature maps to be propagated to the last convolutional layers without any intermediate refinements. 
As such, we are able to more effectively combine features of multiple depths and ERFs. 

The U-Block is defined as follows:
\begin{equation}
    \tilde{\mathbf x}_s =  concat(up(F(\mathbf x_s) + \mathbf x_s), \tilde{\mathbf x}_{s-1})
\end{equation}
where $\mathbf x_s$ denotes the input feature maps, $\tilde{\mathbf x}_{s-1}$ the output feature maps of the same resolution from the previous part, $F(\cdot)$ is the ResBlock's residual mapping (i.e. the convolutional layer stack operations), $up(\cdot)$ is a bilinear upsampling operation that doubles the input's size, and $concat(\cdot, \cdot)$ is simply the concatenation of two feature maps. 

The D-Block is defined as:
\begin{equation}
    \tilde{\mathbf x}_s = concat(down(F(\mathbf x_s) + \mathbf x_s), \tilde{\mathbf x}_{s-1})
\end{equation}
where the notation is the same as before, with $down(\cdot)$ denoting a 2x2 max pooling operation. 

\section{Experimental Setup}

\begin{figure}
\centering
\begin{tabular}{cc}
\subcaptionbox{}{\includegraphics[height=3.4cm,width = 0.22\textwidth]{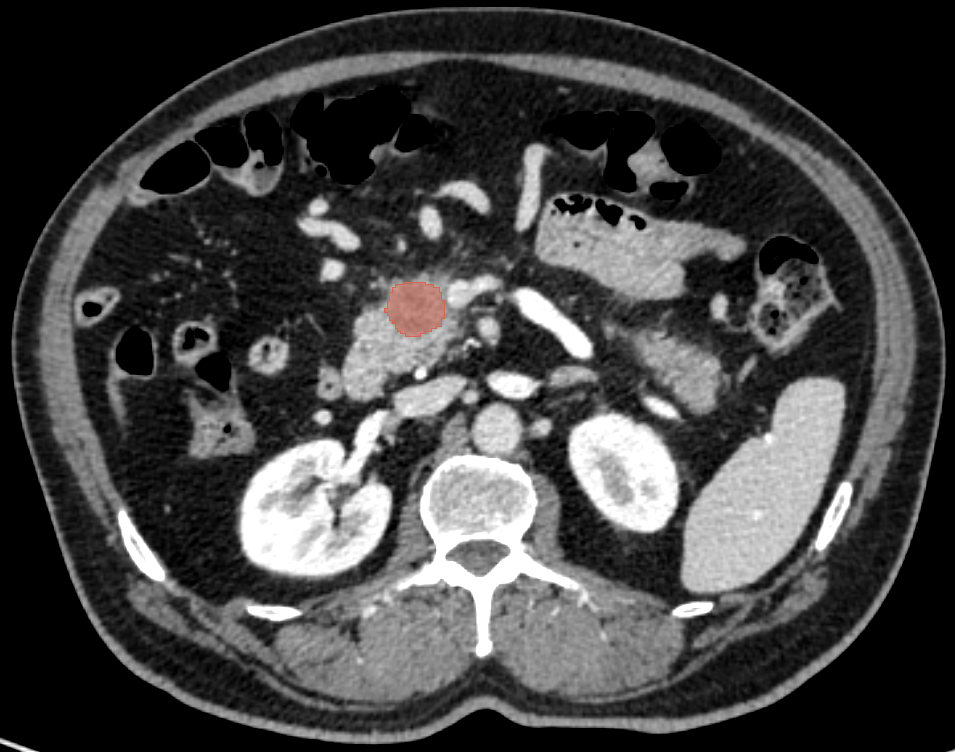}} &
\subcaptionbox{}{\includegraphics[height=3.4cm,width = 0.22\textwidth]{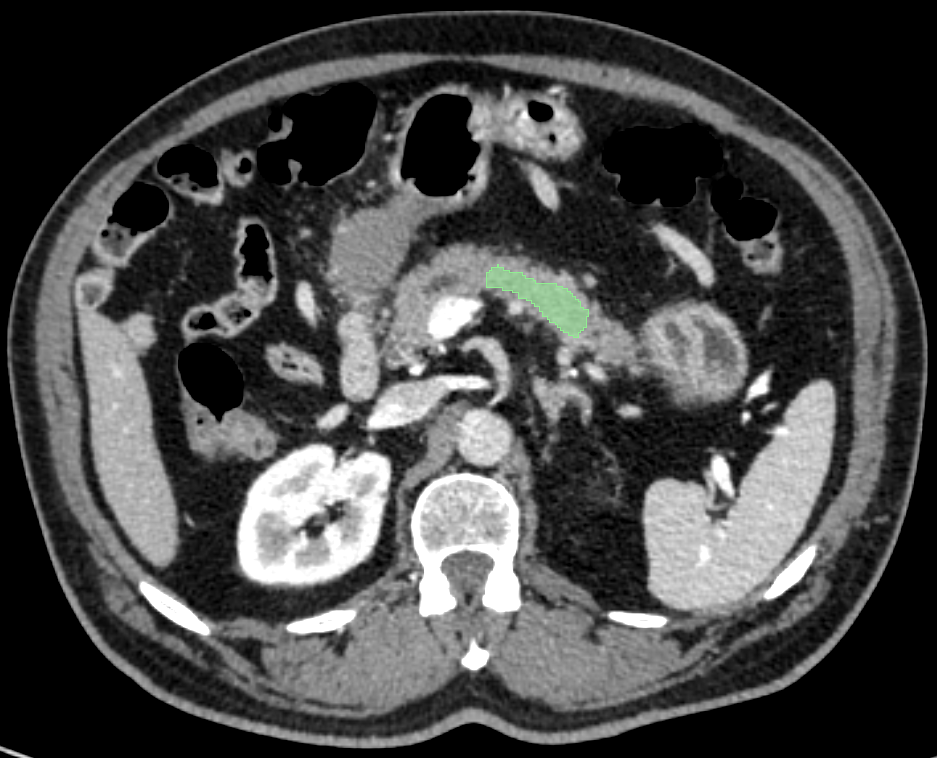}}\\
\end{tabular}
\caption{Annotations of cancerous (red) and non-cancerous (green) areas of the pancreas.}
\label{annotations}
\end{figure}

To train and validate the CEC-CNN, we use a Pancreatic Ductal Adenocarcinoma (PDAC) dataset of 28 patients, containing 1,675 images of very small sizes of varying shapes. 
All patients included in this dataset underwent pancreatectomy for pancreatic cancer in Aretaieion hospital.
The histological findings proved the existence of PDAC in all of them. A pancreatic protocol CT scan was performed pre-operatively in all patients. These CT scans were segmented post-operatively by a senior radiologist specialized in pancreatic imaging, taking into account both surgical and histological findings. Note that these findings cannot accurately be mapped 1:1 to the CT images as the tumors' borders are not always clear due to desmoplastic reaction present around the tumor. Thus, to avoid ``grey areas'' only small patches of definite cancerous and non-cancerous regions were segmented (fig. \ref{annotations}).

To avoid bias based on the label shapes and sizes, we pre-process the images by randomly cropping regions of interest (ROIs) with dimensions higher than 16 pixels to a random size between 8 and 15 pixels. 
We then crop random squares from the ROIs with their dimensions equal to their smallest dimension. 
This pre-processing is re-done on every training epoch, along with a random flip and rotation operation to further augment the dataset. 

To create a benchmark, we train ResNet, MobileNet, and EfficientNet models from scratch on our dataset. To achieve the best possible results we experiment with image input sizes of 32x32, 64x64, by rescaling all images at the network's input using bilinear interpolation.
For networks that benefited from the larger input size, i.e. their $F_1$ score or accuracy increased, we also used an input size of 112x112.
A 70/30 split was chosen for the training and testing data respectively. 
The samples were grouped by the originating patient id, in order to make sure the entirety of the patient's data belonged exclusively to either the train or the test set. 
For the optimizer we chose Stochastic Gradient Descent with an initial learning rate of 0.1 and a momentum of 0.9. 
To deal with class imbalance we computed class weights equal to 0.81 and 1.3 for cancer and non-cancer labels respectively.
These weights were used during training when calculating the Cross-Entropy loss.
The learning rate was reduced by a factor of 0.1 after every 10 epochs with no improvement in the training loss. Each network was trained for 100 epochs.

\section{Results}

Results are shown in table \ref{results}. For each network setup, we calculated the resulting confusion matrix on the same test set. The resulting metrics were calculated as follows:

\begin{equation}
    \text{recall} = \frac{tp}{tp+fn}
\end{equation}

\begin{equation}
    \text{precision} = \frac{tp}{tp+fp}
\end{equation}

\begin{equation}
    F_1 = 2 \times \frac{\text{precision} \times \text{recall}}{\text{precision} + \text{recall}}
\end{equation}

\begin{equation}
    \text{accuracy} = \frac{tp + tn}{tp+tn+fp+fn}
\end{equation}

Where $tp$ is the total count of true positives, $tn$ true negatives, $fp$ false positives, and $fn$ false negatives, with cancer being the positive label, and not-cancer the negative. 

As demonstrated in fig. \ref{annotations}, our dataset annotations come in varied sizes and shapes. Similarly to the training phase, we extract a random square patch from each annotated segment. 
This random process can yield slightly different results in each run; as such we ran the test process 20 times - each time extracting a random square subregion from the same samples - and averaged the resulting metrics to perform a fair comparison.

\begin{table}[htbp]
\caption{Results}
\label{results}
\renewcommand{\arraystretch}{1.25}
\centering
\resizebox{0.48\textwidth}{!}{%
\begin{tabular}{|c|c|r|r|r|r|}
\hline
\textbf{CNN} & \textbf{Input} & \textbf{Recall} & \textbf{Precision} & \textbf{F$_1$} & \textbf{Accuracy} \\
\hline
\multirow{2}{*}
{ResNet18} & 32$\times$32 & 63.5 ±1.3 & 75.4 ±0.9 & 68.9 ±1.0 & 63.6 ±1.1 \\
 & 64$\times$64 & 61.4 ±1.7 & 76.5 ±0.9 & 68.1 ±1.1 & 63.5 ±1.0  \\
\hline
\multirow{3}{*}
{ResNet34} & 32$\times$32 & 63.0 ±1.4 & 71.8 ±0.7 & 67.1 ±1.0 & 60.7 ±0.9 \\
 & 64$\times$64 & 66.7 ±1.4 & 74.6 ±1.2 & 70.4 ±1.2 & 64.4 ±1.3 \\
 & 112$\times$112 & 49.3 ±2.1 & \textbf{79.8 ±1.1} & 60.9 ±1.7 & 59.8 ±1.2 \\
\hline
\multirow{3}{*}
{ResNet50} & 32$\times$32 & 59.4 ±0.9 & 75.5 ±0.9 & 66.5 ±0.7 & 61.9 ±0.8 \\
 & 64$\times$64 & 62.9 ±1.0 & 78.4 ±0.7 & 69.8 ±0.7 & 65.4 ±0.7 \\
 & 112$\times$112 & 59.9 ±1.4 & 78.6 ±1.3 & 68.0 ±1.1 & 64.1 ±1.2\\
\hline
\multirow{2}{*}
{MobileNetv2} & 32$\times$32 & 51.9 ±0.6 & 75.4 ±0.9 & 61.5 ±0.6 & 58.7 ±0.6 \\
 & 64$\times$64 & 67.5 ±1.1 & 74.6 ±0.8 & 70.9 ±0.8 & 64.7 ±0.9 \\
\hline
\multirow{3}{*}
{EfficientNetB0} & 32$\times$32 & 62.8 ±1.3 & 75.2 ±0.9 & 68.5 ±1.0 & 63.2 ±1.1 \\
 & 64$\times$64 & 60.5 ±1.1 & 75.4 ±1.1 & 67.1 ±1.0 & 62.3 ±1.1 \\
 & 112$\times$112 & 65.9 ±1.2 & 74.1 ±0.7 & 69.8 ±0.9 & 63.7 ±0.9\\ 
\hline
\multirow{2}{*}
{EfficientNetB1} & 32$\times$32 & 60.9 ±1.0 & 73.5 ±1.0 & 66.6 ±1.0 & 61.2 ±1.1 \\
 & 64$\times$64 & 60.0 ±1.2 & 74.8 ±0.9 & 66.6 ±0.9 & 61.8 ±0.9 \\
\hline
\multirow{3}{*}
{EfficientNetB2} & 32$\times$32 & 57.8 ±1.3 & 71.5 ±0.9 & 64.0 ±1.0 & 58.6 ±1.0 \\
 & 64$\times$64 & 62.4 ±1.4 & 73.8 ±0.7 & 67.7 ±1.0 & 62.1 ±0.9 \\
  & 112$\times$112 & 63.4 ±1.1 & 75.0 ±0.6 & 68.7 ±0.8 & 63.3 ±0.8 \\
\hline
\multirow{3}{*}
{EfficientNetB3} & 32$\times$32 & 56.6 ±1.7 & 72.9 ±1.1 & 63.7 ±1.4 & 59.0 ±1.3 \\
 & 64$\times$64 & 58.0 ±1.6 & 75.5 ±0.8 & 65.6 ±1.2 & 61.3 ±1.0 \\
 & 112$\times$112 & 64.1 ±0.7 & 75.0 ±0.9 & 69.1 ±0.6 & 63.6 ±0.8 \\
\hline
\multirow{2}{*}
{CEC-CNN} & 32$\times$32 & \textbf{75.5 ±1.1} & 74.8 ±0.7 & \textbf{75.2 ±0.8} & \textbf{68.3 ±0.9} \\
 & 64$\times$64 & 75.3 ±1.5 & 73.0 ±0.7 & 74.1 ±1.0 & 66.6 ±1.1 \\
\hline
\end{tabular}}
\end{table}

Our results show that the CEC-CNN outperforms all other models in all but one metric when the input images are resized to 32$\times$32 pixels.

\begin{figure}
\centering
\begin{tabular}{cccc}
\subcaptionbox{}{\includegraphics[width = 0.09\textwidth]{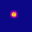}} &
\subcaptionbox{}{\includegraphics[width = 0.09\textwidth]{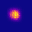}} &
\subcaptionbox{}{\includegraphics[width = 0.09\textwidth]{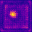}} &
\subcaptionbox{}{\includegraphics[width = 0.09\textwidth]{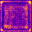}}\\
\end{tabular}
\caption{Visualization of the ERFs of different feature maps at the end of the CEC-CNN, averaged over 100 runs. Multi-sized ERFs can yield multi-scale features.}
\label{erfs}
\end{figure}

Using the CEC-CNN with 32$\times$32 input size, we follow the methodology described in \cite{luoUnderstandingEffectiveReceptive2016} for visualizing the ERF of the final feature maps. 
First, we do a forward pass using a randomly generated image. Then, we set a central pixel of a single feature map equal to 1 and 0 everywhere else, and back-propagate the gradient all the way to the input. The ERF is averaged over 100 runs and then plotted as an image in fig. \ref{erfs}. These visualizations prove that our final layers produce features of varying ERFs.

\section{Conclusion}

In this work we propose a Consecutive Expansion-Contraction CNN (CEC-CNN) architecture that utilizes skip connections and concatenation between same-resolution layers. This allows the network to directly propagate shallow and intermediate feature maps to the network's end, aggregating features of various complexities. In addition, the skip connections allow some feature maps to bypass max-pooling operations, delaying the expansion of their ERFs and allowing them to capture increasingly complex features at small scales. By combining multi-scale features of various depths we are able to more effectively classify small texture images where the size or complexity of their defining features are uncertain. We evaluated our proposed architecture on a dataset of very small patches cropped from pancreatic CT scans, where it outperformed all other state of the art models on the binary classification between cancerous and non-cancerous images.

\bibliographystyle{IEEEtran}
\bibliography{main}

\end{document}